\documentclass[conference]{IEEEtran}
\IEEEoverridecommandlockouts
\usepackage{cite}
\usepackage{amsmath,amssymb,amsfonts}
\usepackage{graphicx}
\usepackage{textcomp}
\usepackage{xcolor}
\usepackage{hyperref}
\usepackage{microtype}
\usepackage{times}
\usepackage{latexsym}
\usepackage{microtype}
\usepackage{graphicx}
\usepackage{multirow}
\usepackage{soul}
\usepackage{amsmath}
\usepackage{amssymb}
\usepackage{enumitem}
\usepackage[normalem]{ulem}
\useunder{\uline}{\ul}{}
\usepackage{setspace}
\usepackage[belowskip=5pt,aboveskip=5pt]{caption}
\renewcommand\vec{\mathbf}
\usepackage{tablefootnote}
\usepackage{authblk}
\usepackage{url}

\usepackage{xspace}
\newcommand{\model}{\texttt{CliniQG4QA}\xspace}

\usepackage{algorithm}
\usepackage{algpseudocode}

\setlist[itemize]{leftmargin=*}

\newcommand{\nop}[1]{}

\def\BibTeX{{\rm B\kern-.05em{\sc i\kern-.025em b}\kern-.08em
    T\kern-.1667em\lower.7ex\hbox{E}\kern-.125emX}}
\begin{document}

\title{CliniQG4QA: Generating Diverse Questions for Domain Adaptation of Clinical Question Answering}

\author[1,*]{Xiang Yue}
\author[1,2,*]{Xinliang Frederick Zhang}
\author[1,3]{Ziyu Yao}
\author[4]{Simon Lin}
\author[1]{Huan Sun}
\affil[1]{The Ohio State University}
\affil[2]{University of Michigan}
\affil[3]{George Mason University}
\affil[4]{Abigail Wexner Research Institute at Nationwide Children’s Hospital}
\affil[ ]{\{\texttt{yue.149, zhang.9975, sun.397\}@osu.edu}}
\affil[ ]{\texttt{ziyuyao@gmu.edu}}
\affil[ ]{\texttt{Simon.Lin@nationwidechildrens.org}}
\affil[*]{These two authors contributed equally}

\maketitle

\begin{abstract}
Clinical question answering (QA) aims to automatically answer questions from medical professionals based on clinical texts. Studies show that neural QA models trained on one corpus may not generalize well to new clinical texts from a different institute or a different patient group, where large-scale QA pairs are not readily available for model retraining.
To address this challenge, we propose a simple yet effective framework, \model, which leverages question generation (QG) to synthesize QA pairs on new clinical contexts and boosts QA models without requiring manual annotations. In order to generate diverse types of questions that are essential for training QA models, we further introduce a seq2seq-based question phrase prediction (QPP) module that can be used together with most existing QG models to diversify the generation. Our comprehensive experiment results show that the QA corpus generated by our framework can improve QA models on the new contexts (up to 8\% absolute gain in terms of Exact Match), and that the QPP module plays a crucial role in achieving the gain.\footnote{Our dataset and code are available at: \href{https://github.com/sunlab-osu/CliniQG4QA/}{https://github.com/sunlab-osu/CliniQG4QA/.}}
\end{abstract}


\section{Introduction}
Clinical question answering (QA), which aims to automatically answer natural language questions based on clinical texts in Electronic Medical Records (EMR), has been identified as an important task to assist clinical practitioners \cite{patrick2012ontology,raghavan2018annotating,pampari2018emrqa,fan2019annotating, rawat2020entity}.  Neural QA models in recent years \cite{chen2017reading, devlin2019bert,rawat2020entity} show promising results in this research. However, answering clinical questions still remains challenging in real-world scenarios, because well-trained QA systems may not generalize well to new clinical contexts from a different institute or a different patient group. For example, as pointed out in \cite{yue2020CliniRC}, when a clinical QA model that was trained on the emrQA dataset \cite{pampari2018emrqa} is deployed to answer questions based on MIMIC-III clinical texts \cite{mimiciii}, its performance drops dramatically by around 30\% even on the questions that are similar to those in training, simply because clinical texts of the two datasets are different (e.g., different topics, note structures, writing styles).


One straightforward solution is to annotate QA pairs on new contexts and retrain a QA model. However, manually creating large-scale QA pairs in clinical domain is extremely challenging due to the requirement of tremendous expert effort, data privacy concerns and other ethical issues.

In this work, we study the problem of \textit{constructing clinical QA models on new contexts without human-annotated QA pairs} (which is referred to as domain adaptation). We assume the availability of a large set of QA pairs on \emph{source} contexts, and our goal is to better answer questions on new documents (\emph{target} contexts\footnote{We use ``new" and ``target" contexts interchangeably.}), where only unlabeled documents are provided.

To this end, we introduce our framework,  \model, which leverages question generation (QG), a recent technique of automatically generating questions from given contexts \cite{du2017learning}, to synthesize clinical QA pairs on target contexts to facilitate the QA model training (Figure~\ref{fig:framework}). The QG model is built up by reusing the QA pairs on source contexts as training data. To apply QG to target contexts, our framework also includes an \textit{answer evidence extractor} (AEE) to extract meaningful text spans, which are worthwhile to {ask questions about}, from the clinical documents. Intrinsically, our framework is backed by the observation that questions in the clinical domain generally follow similar patterns even across different contexts, and clinical QG suffers less from the context shift compared with clinical QA. This allows us to utilize QG models trained on source clinical contexts to boost QA models on target contexts.


However, our preliminary studies find that many existing QG models often fall short on generating questions that are \emph{diverse} enough to serve as useful training data for clinical QA models. To tackle the problem, we introduce a \emph{question phrase prediction} (QPP) module, which takes an answer evidence as input and sequentially predicts potential question phrases (e.g., ``What treatment'', ``How often'') that signify what types of questions humans would likely ask about the answer evidence. By directly forcing a QG model to produce specified question phrases in the beginning of the question generation process (both in training and inference),  QPP enables diverse questions to be generated.

Due to the lack of publicly-available clinical QA pairs for our proposed domain adaptation evaluation setting, we ask clinical experts to annotate a new test set on the sampled MIMIC-III \cite{mimiciii} clinical texts. We conduct extensive experiments to evaluate \model, using emrQA \cite{pampari2018emrqa} as the source contexts and our annotated MIMIC-III \cite{mimiciii} as the target ones. We instantiate our framework with a variety of widely adopted base QG models and base QA models.

By performing comprehensive analyses, we show that the proposed QPP module can substantially help generate much more diverse types of questions (e.g., ``When'' and ``Why'' questions). More importantly, we systematically demonstrate the strong capability of \model for improving QA performance on new contexts by evaluating it on our constructed MIMIC-III QA dataset. 
When using QA pairs automatically synthesized by our QPP-enhanced QG models as the training corpus, we are able to boost QA models' performance by up to 8\% in terms of Exact Match (EM), compared with their counterparts directly trained on the emrQA dataset. To further investigate why QG boosts QA, we provide both quantitative and qualitative analyses, indicating that QA models can benefit from seeing more target contexts as well as more diverse questions generated on them.

\section{Preliminary and Related Work}
\begin{figure}[t]
    \centering
    \includegraphics[width=\linewidth]{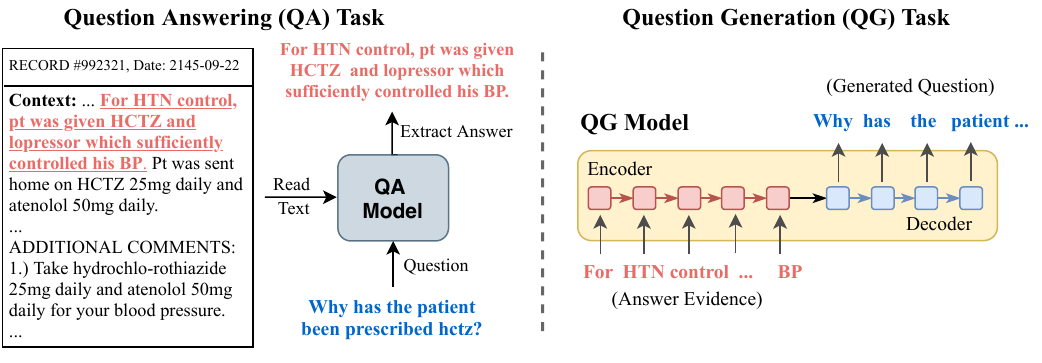}
    \caption{Illustration of Clinical Question Answering (QA) and Question Generation (QG) task. }
    \label{fig:QA_QG_task}
    \vspace{-15pt}
\end{figure}

{\noindent\textbf{Clinical Question Answering} aims to
extract a text span (a sentence or multiple sentences) as the answer from a patient clinical note given a question (Fig. \ref{fig:QA_QG_task} left) \cite{yue2020CliniRC}. 
Though many neural models \cite{Seo2017Bidirectional,chen2017reading, devlin2019bert,rawat2020entity,wen2020adapting} have achieved impressive results on this task, their performance on new clinical contexts, whose data distributions could be different from the ones that these models were trained on, is still far from satisfactory \cite{yue2020CliniRC}. Though one can improve the performance by adding more QA pairs on new contexts into training, however, manually creating large-scale QA pairs in the clinical domain often involves tremendous expert effort and data privacy concerns.}
Moreover, during the pandemic, clinical QA models can also be deployed to answer COVID-19 related questions \cite{poliak,zhang2021cough}.


\noindent\textbf{Question Generation} seeks to automatically generate questions given a sentence or paragraph (Fig. \ref{fig:QA_QG_task} right). Existing QG models \cite{du2017learning, zhou2017neural, sun2018answer, zhao2018paragraph, nema2019let, tuan2020capturing, yang2017semi, du2018harvesting, alberti2019synthetic, zhang2019addressing} in the open domain usually adopt a seq2seq (encoder-decoder) architecture. One of the drawback of such models is that they can only generate one question given one input and fail to generate multiple diverse questions, which we find is crucial to the QA task. Some recent work \cite{kang2019let,cho2019mixture,liu2020asking} explores the diverse QG in the open domain, but they cannot be directly applied to the clinical domain as their models usually require a short answer (e.g., an entity) as input but that information sometimes is not available in the clinical QA dataset (e.g. emrQA \cite{pampari2018emrqa}), rendering the difficulty of directly deploying their model on the clinical QA.

In the clinical and medical domain, \cite{shen2020generation} and \cite{soni2019paraphrase,soni2020paraphrasing} apply Variational Autoencoder (VAE) models to generate or paraphrase medical or clinical questions. However, none of them explore leveraging QG to improve QA performance on new contexts.

\noindent \textbf{Our aim} is to improve clinical QA on new clinical texts (i.e., domain adaptation of clinical QA). We assume the availability of a large set of QA pairs and corresponding clinical documents (source contexts), and our goal is to better answer questions on new documents (target contexts) where only unlabeled documents are provided. We leverage a QG model to synthesize diverse QA pairs to save medical experts annotation efforts and improve QA performance without requiring extra annotations. Our setting is very practical in the real-world scenario, since it is infeasible to always annotate QA pairs on new clinical texts when deploying a QA system into a new environment.

\section{Methods}
\subsection{Overview of Our Framework}

\begin{figure*}[t]
    \centering
    \includegraphics[width=0.9\linewidth]{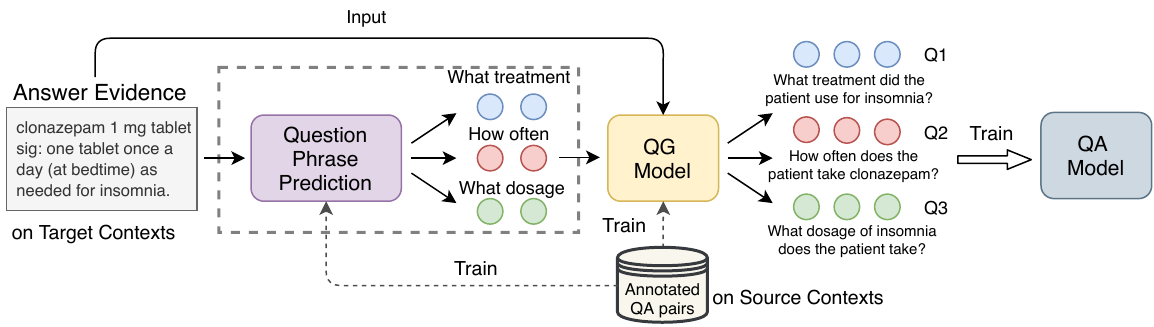}
    \caption{Illustration of our Question Phrase Prediction (QPP) module, which can be used together with QG models to diversify generations.}
    \label{fig:framework}
    \vspace{-15pt}
\end{figure*}

We first give an overview of our \model framework (Fig~\ref{fig:framework}). \model improves clinical QA on new contexts by automatically synthesizing QA pairs for new clinical contexts. To approach this, we first leverage an \textit{answer evidence extractor} to extract meaningful text spans from unlabeled documents, based on which a QG model can be applied to generate questions.

In order to encourage diverse questions, we reformulate the question generation process as two-stage. In the first stage, we propose a \textit{question phrase prediction} module to predict a set of question phrases, which represent the types of questions humans would ask, given an answer evidence. In the second stage, following a specific question phrase predicted by our QPP, a QG model is used to complete the rest of the question.

Therefore, our framework \model is able to produce questions of more diverse types. The generated QA pairs by QG models are finally used to train QA models on new contexts.


\vspace{1pt}
\subsection{Answer Evidence Extractor (AEE)}
\label{sec:ans_extract}
When human annotators create questions, they first read a document and then select a text span to ask questions about. To imitate this process, we implement an \textit{answer evidence extractor} to extract possible text spans from a document. Following \cite{pampari2018emrqa,yue2020CliniRC}, we focus on longer text spans (as answer evidences) instead of short answers (e.g., a single named entity), since longer text spans often contain  richer information compared with short ones, which are very important for the clinical QA task.  

More formally, given a document (context) $\vec{p}=\{p_1,p_2,...,p_m\}$, where $p_i$ is the $i$-th token of the document and $m$ is the total number of tokens, we aim to extract potential evidence sequences. Since the answer evidence is not always a single sentence (sometimes could be multiple sentences), instead of treating it as a sentence selection task, we formulate it as a sequence labeling (or tagging) task. We follow the \texttt{BIO} tagging (short for beginning, inside, outside), a commonly used sequence labeling scheme \cite{ramshaw1999text}, to label answer evidences.

Firstly, we adopt the ClinicalBERT model \cite{alsentzer2019publicly}
to encode the document:
\vspace{-6pt}
\begin{equation}
\vspace{-2pt}
    \mathbf{U} = \texttt{ClinicalBERT} \{p_1,...,p_m\}.
\end{equation}
where $\mathbf{U} \in \mathbb{R}^{m\times d}$, and $d$ is size of the dimension. 

Following the same paradigm of the BERT model for the sequence labeling task \cite{devlin2019bert}, we use a linear layer on top of the hidden states output by ClinicalBERT followed by a softmax function to do the classification:
\vspace{-5pt}
\begin{equation}
\vspace{-3pt}
    \Pr (a_j|p_i)=\text{softmax}(\mathbf{U} \cdot \mathbf{W} + \mathbf{b}), \;\forall p_i\in \vec{p}
\end{equation}
where $a_j$ is the predicted \texttt{BIO} tag. 


After prediction, we observe that the extracted answer evidences sometimes are broken sentences due to the noisy nature and uninformative language (e.g., acronyms) of clinical texts. To make sure that the extracted evidences are meaningful, we designed a \textit{``merge-and-drop''} heuristic rule to further improve the extractor’s accuracy. Specifically, for each extracted evidence candidate, we first examine the \textit{length}  (number of tokens) of the extracted evidence. If the length is larger than the threshold $\eta$, we keep this evidence; otherwise, we compute the \textit{distance}, i.e., the number of tokens between the current candidate span and another closest candidate span If the \textit{distance} is smaller than the threshold $\gamma$, we merge these two ``close-sitting'' spans; otherwise, we drop this overly-short evidence span. In our experiments, we set $\eta$ and $\gamma$ to be 3 and 3, respectively, since they help achieve the best performance on the dev set.

\subsection{Question Phrase Prediction (QPP)}
\label{qpp}
Existing QG models are often biased to generate limited types of questions. To address this problem, we introduce our question phrase prediction module that can be used to diversify the generation of existing QG models. 


{Formally, denote $V_l=\{s_1, ..., s_L\}$ as the vocabulary of all available question phrases of length $l$ in the training data and $L=|V_l|$ as its size. $V_l$ can be obtained by collecting the first $n$-gram words in the questions. We set $n=2$ in our experiment as it achieves the best performance on the dev set. Given an answer evidence $\vec{a}$, the goal of QPP is to map $\vec{a} \rightarrow \vec{y}=(y_1, ..., y_L) \in \{0, 1\}^L$, where $y_i=1$ indicates predicting $s_i$ in $V_l$ as a question phrase for the evidence $\vec{a}$.  Instead of treating it as a common multi-label classification problem, we formulate the task as \emph{a sequence prediction} problem and adopt a commonly used seq2seq model with an attention mechanism \cite{luong2015effective} to predict a sequence of question phrases $\vec{s}=(s_{j_1}, ..., {s_{j_{|\vec{s}|}}})$ (e.g., ``What treatment'' ($s_{j_1}$) $\rightarrow$ ``How often'' ($s_{j_2}$) $\rightarrow$ ``What dosage'' ($s_{j_3}$), with $|\vec{s}|=3$).}



During training, we assume that the set of question phrases is arranged in a pre-defined order. Such orderings can be obtained with some heuristic methods, e.g., using a descending order based on question phrase frequency in the corpus\footnote{In our dataset, each answer evidence is tied with multiple questions, which allows the training for QPP.}.
In the inference stage, QPP can dynamically decide the number of question phrases for each answer evidence by predicting a special \texttt{[STOP]} token.
By decomposing QG into two steps (diversification followed by generation), the implemented QPP can increase the diversity in a more controllable way.

\newcommand\Algphase[1]{%
\vspace*{-.7\baselineskip}\Statex\hspace*{\dimexpr-\algorithmicindent-0pt\relax}\rule{1.08\linewidth}{0.6pt}%
\vspace{-4pt}
\Statex\hspace*{-\algorithmicindent}\textbf{#1}%
\vspace*{-.7 \baselineskip}\Statex\hspace*{\dimexpr-\algorithmicindent-0pt\relax}\rule{1.08\linewidth}{0.6pt}%
}
\setlength{\textfloatsep}{20pt}%

\begin{algorithm}[t]
 \caption{\model training procedure}
 \label{alg:model}
 \textbf{Input}: labeled \textit{source} data $\{(P_S,A_S,Q_S)\}$, unlabeled \textit{target} data $\{P_T\}$ \\
 \textbf{Output}: Generated QA pairs $\{(A'_T,Q'_T)\}$  on \textit{target} contexts; An optimized QA model for answering questions on target contexts;
 \begin{algorithmic}[1]
 \Algphase{Pretraining Stage}

\State Train \textit{Answer Evidence Extractor} based on the \textit{source} data $\{(P_S,A_S)\}$ using Eq. \ref{eq:Loss_AE}

\State Obtain question phrase data $Y_S$ from $Q_S$ and  train \textit{Question Phrase Prediction} module on the \textit{source} data $\{(A_S, Y_S)\}$ using Eq. \ref{eq:Loss_QPP}

\State Train a \textit{QPP-enhanced QG} model on the \textit{source} data $\{(A_S, Y_S,Q_S)\}$ using Eq. \ref{eq:Loss_QG}

\Algphase{Training Stage}

\State Use \textit{AEE} to extract potential answer evidences $\{A'_T\}$ on the \textit{target} contexts $\{P_T\}$

\State Use \textit{QPP} to predict potential question phrases set $\{Y'_T\}$ on $\{A'_T\}$ 

\State Use \textit{QPP-enhanced QG} to generate diverse questions $\{Q'_T\}$ based on  $\{(A'_T,Y'_T)\}$

\State Train a \textit{QA} model on synthetic \textit{target} data $\{(P_T, A'_T,Q'_T)\}$  using Eq. \ref{eq:Loss_QA}
\end{algorithmic} 
\end{algorithm}

\subsection{Training}
Algorithm \ref{alg:model} illustrates the pretraining and training procedure of our \model.

During the \textit{pretraining} stage, we first train the answer evidence extractor (AEE) module on the source contexts by minimizing the negative log-likelihood loss:
\vspace{-4pt}
\begin{equation}
\vspace{-4pt}
\label{eq:Loss_AE}
    {L}_{AEE}=-\sum_{i}\log P (\vec{a}|\vec{p};\phi)
\end{equation}
where $\phi$ represents all the parameters of the answer evidence extractor. For the supervision signals, we identify all evidences in the source data as ground-truth chunks which are marked using the \texttt{BIO} scheme.


Moving to the Question Phrase Prediction (QPP) module, given an answer evidence $\vec{a}$, we aim to predict a question phrase sequence $\vec{y}$ and minimize:
\vspace{-4pt}
\begin{equation}
\vspace{-4pt}
\label{eq:Loss_QPP}
    {L}_{QPP}=-\sum_{i} \log P (\vec{y}|\vec{a};\theta)
\end{equation}
where $\theta$ denotes all the parameters of QPP.

Then we can train any QG model (e.g, NQG \cite{du2017learning}) on source data by minimizing:
\vspace{-4pt}
\begin{equation}
\vspace{-4pt}
\label{eq:Loss_QG}
    {L}_{QG}=-\sum_{i} \log P (\vec{q}|\vec{a}, \vec{y};\mu)
\end{equation}
where $\mu$ denotes all parameters of the QG model. 

During the \textit{training} stage, given unlabeled target clinical documents, we first extract answer evidences, based on which QPP can be ``plugged" into the QG model to generate diverse questions. Finally, a QA model (e.g., DocReader \cite{chen2017reading}) can be trained on the generated QA pairs of the target documents: 
\vspace{-4pt}
\begin{equation}
\vspace{-4pt}
\label{eq:Loss_QA}
    {L}_{QA}=-\sum_{i} \log P (\vec{a}|\vec{q}, \vec{p};\delta)
\end{equation}
where $\delta$ denotes all parameters of the QA model.









\begin{table}[t]
\centering
\caption{Statistics of the datasets. We synthesize a machine-generated dev set and ask human experts to annotate a test set for MIMIC-III.}
\label{tbl:dataset}
\begin{tabular}{ccc}
\hline
 (Question / Context) & emrQA & MIMIC-III \\ \hline
\# Train  & 781,857 / 337 & - / 337 \\
\# Dev  & 86,663 / 41 & 8,824 / 40 \\
\# Test & 98,994 / 42 & 1,287 / 36 \\
\# Total & 967,514 / 420 & - / 413 \\ \hline
for purpose of & \begin{tabular}[c]{@{}c@{}}QG \& QA \\ (source)\end{tabular} & \begin{tabular}[c]{@{}c@{}}QA\\ (target)\end{tabular} \\ \hline
\end{tabular}%
\end{table}

\section{Generalizability Test Set Construction}
Unlike open domain, there are very few publicly available QA datasets in the clinical domain. EmrQA dataset \cite{pampari2018emrqa}, which was generated based on medical expert-made question templates and existing annotations on n2c2 challenge datasets \cite{n2c2}, is a commonly adopted dataset for clinical reading comprehension. 

However, all the QA pairs in emrQA are based on n2c2 clinical texts and thus not suitable for our generalization setting. \cite{yue2020CliniRC} studied a similar problem and annotated a test set on MIMIC-III clinical texts \cite{mimiciii}. However, their test set is too small (only 50 QA pairs) and not publicly available. Given the lack of a reasonably large clinical QA test set for studying generalization, with the help of three clinical experts, we create 1287 QA pairs on a sampled set of MIMIC-III \cite{mimiciii} clinical notes, \textit{which have been reviewed and approved by PhysioNet\footnote{\url{https://physionet.org/}. PhysioNet is a resource center with missions to conduct and catalyze for biomedical research, which offers free access to large collections of physiological and clinical data, such as MIMIC-III \cite{mimiciii}.} and is downloadable by following the instructions\footnote{https://physionet.org/content/mimic-iii-question-answer/1.0.0/.}}.
\begin{table*}[t]
\centering
\caption{The QA performance on MIMIC-III test set. emrQA is also included as a baseline dataset to help illustrate the generated diverse questions on MIMIC-III are useful to improve the QA model performance on new contexts.}
\label{tbl:qa_results}
\begin{tabular}{l|cccccc||cccccc}
\hline
\multirow{4}{*}{\textbf{QA Datasets}} & \multicolumn{6}{c||}{\textbf{DocReader} \cite{chen2017reading}} & \multicolumn{6}{c}{\textbf{ClinicalBERT} \cite{alsentzer2019publicly}} \\ \cline{2-13} 
 & \multicolumn{2}{c|}{\begin{tabular}[c]{@{}c@{}} \textbf{Human}\\  \textbf{Generated} \end{tabular}} & \multicolumn{2}{c|}{{\begin{tabular}[c]{@{}c@{}} \textbf{Human}\\  \textbf{Verified} \end{tabular}} } & \multicolumn{2}{c||}{{\begin{tabular}[c]{@{}c@{}} \textbf{Overall}\\  \textbf{Test} \end{tabular}} } & \multicolumn{2}{c|}{\begin{tabular}[c]{@{}c@{}} \textbf{Human}\\  \textbf{Generated} \end{tabular}} & \multicolumn{2}{c|}{{\begin{tabular}[c]{@{}c@{}} \textbf{Human}\\  \textbf{Verified} \end{tabular}} } & \multicolumn{2}{c}{{\begin{tabular}[c]{@{}c@{}} \textbf{Overall}\\  \textbf{Test} \end{tabular}} } \\ \cline{2-13} 
 & \textbf{EM} & \textbf{F1} & \textbf{EM} & \textbf{F1} & \textbf{EM} & \textbf{F1} & \textbf{EM} & \textbf{F1} & \textbf{EM} & \textbf{F1} & \textbf{EM} & \textbf{F1}  \\ \hline
 {\begin{tabular}[l]{@{}l@{}} emrQA \cite{pampari2018emrqa} \end{tabular}} 
  & 69.87 & 83.66 & 61.44 & 78.82 & 63.48 & 79.99 & 69.23 & 82.83 & 61.23 & 78.56 & 63.17 & 79.59  \\ \hline\hline
 {\begin{tabular}[c]{@{}c@{}} NQG \cite{du2017learning}\\      \end{tabular}} 
 & 66.99 & 79.67 & 64.71 & 79.36 & 65.26 & 79.43 & 67.30 & 82.59 & 59.49 & 76.68 & 61.38 & 78.11 \\
+ BeamSearch & 71.15 & 83.07 & 67.07 & 81.21 & 68.07 & 81.66 & 68.91 & 84.26 & 63.17 & 79.17 & 64.56 & 80.40 \\
+ Top-k Sampling & 71.58 & 83.48 & 66.77 & 80.45 & 67.94 & 81.19 & 67.74 & 81.96 & 60.82 & 78.16 & 62.50 & 79.08 \\
+ Nucleus Sampling & 70.62 & 83.68 & 67.16 & 80.37 & 68.00 & 81.17 & 68.70 & 83.21 & 62.36 & 77.89 & 63.90 & 79.18 \\
\textbf{+ QPP (Ours)} & \textbf{74.36} & \textbf{85.18} & \textbf{68.82} & \textbf{82.89} & \textbf{70.09} & \textbf{83.44} & \textbf{69.23} & \textbf{84.33} & \textbf{63.79} & \textbf{79.56} & \textbf{65.11} & \textbf{80.72} \\ \hline\hline
 {\begin{tabular}[c]{@{}c@{}} NQG++ \cite{zhou2017neural}  \end{tabular}} & 66.34 & 81.34 & 65.94 & 78.71 & 66.04 & 79.35 & 65.06 & 80.11 & 59.59 & 75.85 & 60.92 & 76.88 \\
+ BeamSearch & 72.11 & 84.56 & 68.10 & 80.09 & 69.07 & 81.17 & 68.26 & 83.70 & 64.61 & 80.30 & 65.50 & 81.12 \\
+ Top-k Sampling & 73.29 & 85.56 & 69.11 & 82.38 & 69.41 & 83.35 & 70.19 & 85.61 & 62.84 & 79.77 & 64.62& 81.19 \\
+ Nucleus Sampling &73.34 & 84.95 & 68.94 & 81.72 & 70.01 & 82.51 & 70.19 & 84.72 & 63.93 & 79.54 & 65.45 & 80.80 \\
\textbf{+ QPP (Ours)} & \textbf{74.68} & \textbf{85.92} & \textbf{70.05} & \textbf{83.47} & \textbf{71.10} & \textbf{84.06} & \textbf{70.83} & \textbf{85.76} & \textbf{65.33} & \textbf{80.64} & \textbf{66.67} & \textbf{81.88} \\ \hline\hline
{\begin{tabular}[c]{@{}c@{}} BERT-SQG \cite{chan2019recurrent}  \end{tabular}}
 & 70.19 & 81.47 & 66.05 & 79.64 & 67.05 & 80.08 & 65.06 & 82.20 & 59.59 & 78.04 & 60.92 & 79.05 \\
+ BeamSearch & 73.71 & 84.44 & 68.71 & 81.98 & 69.93 & 82.58 & 67.31 & 82.54 & 61.94 & 79.02 & 63.25 & 79.88 \\
+ Top-k Sampling & 72.81 & 84.16 & 69.20 & 82.24 & 70.07 & 82.71 & 69.12 & 84.20 & 60.44 & 78.27 & 62.55 & 79.71 \\
+ Nucleus Sampling & 70.73 & 83.60 & 68.56 & 81.80 & 69.09 & 82.24 & 67.74 & 83.16 & 61.61 & 78.74 & 63.09 & 79.81 \\
\textbf{+ QPP (Ours)} &\textbf{74.36} & \textbf{85.53} & \textbf{70.77} & \textbf{83.60} & \textbf{71.64} & \textbf{84.07} & \textbf{69.23} & \textbf{85.38} & \textbf{64.21} & \textbf{80.53} & \textbf{65.43} & \textbf{81.71}\\ \hline
\end{tabular}%
\end{table*}

\noindent\textbf{Annotation Process.} 
We sample 36 MIMIC-III clinical notes as contexts. When sampling MIMIC-III notes, we ensure that all the sampled clinical texts do not appear in emrQA, acknowledging that there is a small overlap between the two datasets. For each context, clinical experts can ask any questions as long as an answer can be extracted from the context. To save annotation effort, QA pairs generated by QG models (i.e., all base QG models and their diversity-enhanced variants; see Section \ref{ref:base}) are provided as references, and duplicates are removed. Meanwhile,  clinical experts are \textit{highly encouraged} to create new questions based on the given clinical text (which are marked as \textit{``human-generated"/``HG''}). But if they do find the machine-generated questions sound natural and match the provided answer, they can keep them (which are marked as \textit{``human-verified"/``HV''}). After obtaining the annotated questions, we ask another clinical expert to do a final pass of the questions in order to further ensure the quality of the test set.  The final test set consists of 1287 questions (of which 975 are \textit{``human-verified"} and 312 are \textit{``human-generated"}). 

We understand that there might be potential bias when evaluating QA models on the HV set (i.e, a QG model which is used to generate training questions for a QA model also contributes questions to the HV set as well). However, such bias might exist in human annotated data as well (e.g., the same set of humans create both training and testing dataset). Note that the contexts used to generate questions in HV/HG are separated from those to generate training questions. Besides, due to the relatively limited language patterns in clinical domain, we find most questions in HV set sound like what humans would ask. As such, we still deem it as a valuable asset and potential future research could leverage our HV set as their dev set to tune hyper-parameters.

To help tune the model, we also construct dev set of MIMIC-III  by sampling generated questions from QG models and their variants and is used to tune the hyper-parameters. 
In the following sections, we consider emrQA as the \textit{source} dataset and our annotated MIMIC-III QA dataset as the \textit{target} data. Detailed statistics of the two datasets are in Table \ref{tbl:dataset}.




\section{Experimental Setup}
\vspace{-2pt}

\vspace{-2pt}
\subsection{Base QG models}
\label{ref:base}
We instantiate our \model framework using three base QG models:

\noindent $\bullet$  \textbf{NQG} \cite{du2017learning} is the first seq2seq model with a global attention mechanism \cite{luong2015effective} for question generation. 

\noindent $\bullet$ \textbf{NQG++} \cite{zhou2017neural} is one of the most commonly adopted QG baselines with a feature-enriched encoder (e.g., lexical features) and a copy mechanism \cite{gulcehre2016pointing}.

\noindent $\bullet$ \textbf{BERT-SQG} \cite{chan2019recurrent} uses a pretrained BERT model (we use ClinicalBERT \cite{alsentzer2019publicly} to accommodate clinical setting) as the encoder and formulates the decoding as a ``MASK" token prediction problem.


It has been studied that beam search and sampling strategies show  competitive  performance in diversifying generations \cite{IppolitoKSKC19,diversity2020arafat}. We thus include Top-k \cite{topk} and Nucleus samplings \cite{holtzman2019curious} as representative sampling strategies in our experiments.

As such, to investigate the effectiveness of diverse QG for QA, we consider the following variants of each base QG model: (1) Base Model: Inference with greedy search; (2) Base Model + Beam Search: Inference with Beam Search of beam size $K$ and keep top $K$ beams ($K=3$); (3) Base Model + Top-k sampling: Inference with sampling from top-k tokens ($k=20$); (4) Base Model + Nucleus sampling: Inference with sampling from top-p tokens ($p=0.95$); (5) Base Model + QPP: Inference with greedy search for both QPP module and Base model.


\subsection{Base QA models}
For QA, we instantiate \model with two base models, DocReader \cite{chen2017reading} and ClinicalBERT \cite{alsentzer2019publicly}. When training a QA model, we only use the synthetic data on the target contexts and do not combine the synthetic data with the source data since the combination does not help in our preliminary studies.

Note that more complex QG/QA models and training strategies can also be used in our framework. As this work focuses on exploring how \emph{diverse} questions help QA on target contexts, we adopt fundamental QG/QA models and training strategies, and leave more advanced ones that are complementary to our framework as future work.

\subsection{Evaluation Metrics}
\label{eval_metric}




For QA evaluation, we report exact match (EM) (percentage of predictions that match the ground truth answers exactly) and F1 (average overlap between the predictions and
ground truth answers) as in \cite{rajpurkar2016squad}. Since our main goal is to evaluate whether the generated questions are useful to improve the QA performance on the target contexts, the common language generation metrics such as BLEU \cite{papineni2002bleu} and ROUGE-L \cite{lin-2004-rouge} are not suitable to reflect the quality of the generated questions, and thus we do not adopt these metrics in our experiments.


\subsection{Implementation Details}
\noindent\textbf{Base QG Models:} We re-implement three base QG models using Pytorch and have ensured that they achieve comparable performance as originally reported. Best QG models are selected using the per-token accuracy of both the QPP module (if applicable) and QG on dev set.

\noindent\textbf{Base QA Models:} We use the open-sourced implementation.\footnote{DocReader: https://github.com/facebookresearch/DrQA.
ClinicalBERT: https://github.com/EmilyAlsentzer/clinicalBERT.} Best QA models are selected using EM and F1 on dev set. 

\noindent\textbf{Hyperparameters Search:} Hyperparameters of QG models are set to be the same as in original papers and hyperparameters of QA models are set according to \cite{yue2020CliniRC}. Specifically, we train NQG and NQG++ up to 20 epochs, BERT-SQG up to 5 epochs, DocReader up to 5 epochs and ClinicalBERT up to 3 epochs.


\section{Experimental Results}

\begin{figure*}[t]
    \centering
    \includegraphics[width=\linewidth]{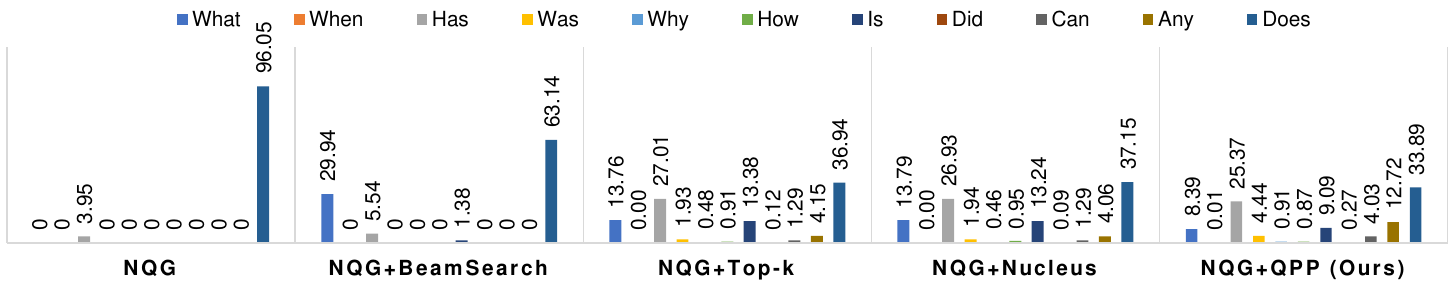}
    \caption{Distributions over types of questions generated by NQG models.}
    \label{fig:ques_distibution}
    \vspace{-15pt}
\end{figure*}

\subsection{Can Generated Questions Help QA on New Contexts?} 

Table \ref{tbl:qa_results} summarizes the performance of two widely used QA models, DocReader \cite{chen2017reading} and ClinicalBERT \cite{alsentzer2019publicly}, on the MIMIC-III testing set. The QA models are trained based on different corpora, including the emrQA dataset as well as QA pairs generated by different models.  For a fair comparison, we keep the total number of generated QA pairs roughly the same as emrQA.
As can be seen from the table, the QA models based on the corpora that are generated using the three base QG models can only achieve roughly the same or even worse performance compared with the QA models trained on the emrQA dataset. Though the Beam Search and sampling strategies could boost the diversity of generated questions to some extent, and thus lead to the improvement of QA models, our proposed QPP module can improve the QA performance by a larger margin.  For example, training DocReader using questions generated by NQG++ with our QPP module outperforms that using the emrQA dataset by around 8\% under EM and 4\% under F1 on the overall test set. Moreover, the results on human-generated portion are consistently better than that on human-verified. It's attributed to the fact that human-created questions are more readable and sensible while human-verified questions are a bit of less natural though correctness is ensured.

All these results indicate that generating a diverse QA corpus is useful for downstream QA on new contexts, and our simple QPP module can help existing QG models achieve such a goal.

\subsection{Why QG Boosts QA on New Contexts?}
To further explore why QG can boost QA, we consider three major factors when generating a QA corpus: the number of documents, the number of answer evidences per document, and the number of generated questions per answer evidence. When we test one factor, we fix the other two. For example, we fix the number of answer evidences and questions at 20 and 6 when we test the influence of the number of documents. We use NQG++ and DocReader as our base QG and QA models to instantiate our \model framework and report the performance on the Dev set.

\begin{figure}[t]
    \centering
    \includegraphics[width=\linewidth]{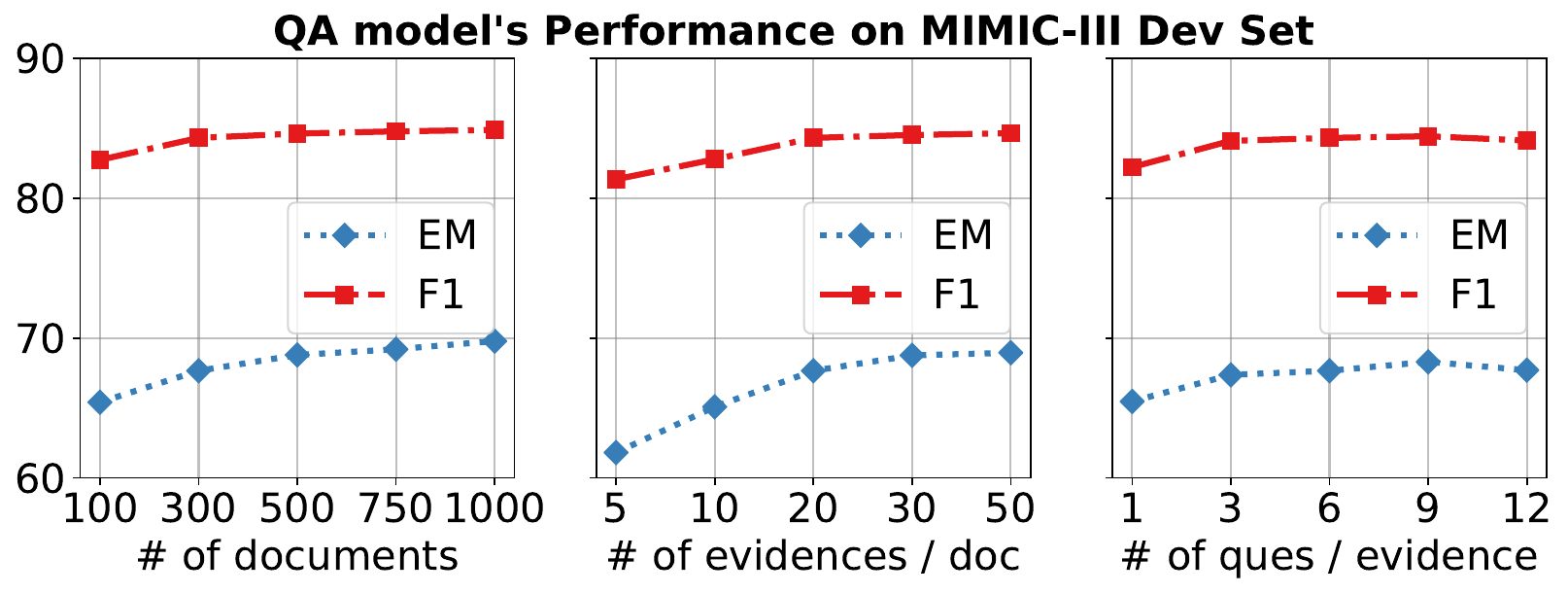}
    \caption{Influence of the number of documents, number of evidences per document, number of QA pairs per evidence on QA performance.}
    \label{fig:qa_influence}
\end{figure}

As can seen from Fig \ref{fig:qa_influence}, the performance steadily increases when we use more documents and more answer evidences during QA corpus generation. This can demonstrate the first hypothesis: The generated corpus enables a QA model to see more new contexts during training, which can help the QA model get a better understanding of similar contexts during testing. The more contexts it sees, the more benefits it could obtain. 
We can also see that with the increase of the number of generated questions per evidence, the performance 
generally rises up. This indicates that multiple diverse questions are essential for boosting QA performance.

\noindent\textbf{A Closer Look at Generated Question Types.}
\label{sec:distribution_question}
To further demonstrate QPP module can help generate diverse questions, we show the distribution over the types of questions generated by NQG-based models in Fig \ref{fig:ques_distibution}. 



We observe that questions generated by base NQG and NQG+BeamSearch are limited in terms of the question types. However, more types of questions (e.g., ``How", ``Why") can be generated when enabling sampling strategies. Furthermore, when being equipped with our QPP module, the NQG model can even generate questions of an extremely rare type, i.e., "When" questions. Though Top-k and Nucleus sampling methods also generate questions of less frequent types, our QPP module could cover even more types. 

In summary, we think seeing many new contexts and diverse questions are the two main reasons why QA models are boosted.
\begin{figure}[t]
    \centering
    \includegraphics[width=\linewidth]{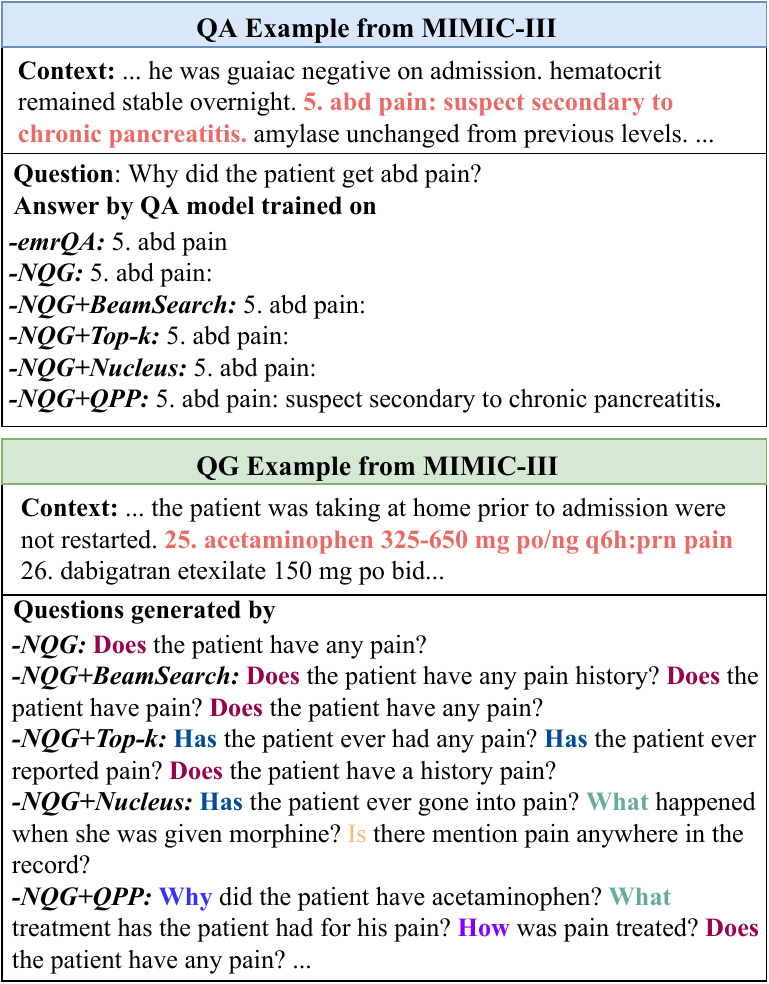}
    \caption{QA and QG examples. The red parts in contexts are ground-truth answer evidences. } 
    \label{fig:qualitative}
\end{figure}

\subsection{Diverse Questions Really Matter for QA: Two Real Cases.} In Fig \ref{fig:qualitative}, we present a QA example and a QG example from MIMIC-III for qualitative analysis.

In the QA example, this ``why" question can be correctly answered  by the QA model (DocReader) trained on the ``NQG+QPP" generated corpus while the QA models trained on other generated corpora fail. This is because, as shown in Fig \ref{fig:ques_distibution}, the NQG model and ``NQG+BeamSearch" cannot generate any ``why" questions and sampling strategies could only help generate a limited number of ``why'' questions. Thus QA models trained on such corpora cannot answer questions of less frequent types. Though the emrQA dataset contains diverse questions (including ``why" questions),  its contexts might be different from MIMIC-III in terms of topic, note structures, writing styles, etc.  So the model trained on emrQA struggles to answer some questions as well.

In the QG example, the base model NQG can only generate one question. Though utilizing the Beam Search enables the model to explore multiple candidates, the generated questions are quite similar and are less likely to help improve QA. Sampling strategies, though further diversifying the generation during decoding, suffer from generating irrelevant contents (e.g., ``NQG+Nucleus'' generates a irrelevant ``morphine'' token). Enabling our QPP module helps generate relevant and diverse questions including ``Why", ``What", ``How", etc.

\subsection{Ablation Study}
\label{ref:as}
\begin{table}[t]
\centering
\caption{The QA performance on MIMIC-III test set when QPP is employed with sampling strategies}
\label{tbl:qa_results_side}
\resizebox{\linewidth}{!}{%
\begin{tabular}{l|cccccc}
\hline
\multirow{4}{*}{\textbf{QA Datasets}} & \multicolumn{6}{c}{\textbf{DocReader} \cite{chen2017reading}}  \\ \cline{2-7} 
 & \multicolumn{2}{c|}{\begin{tabular}[c]{@{}c@{}} \textbf{Human}\\  \textbf{Generated} \end{tabular}} & \multicolumn{2}{c|}{{\begin{tabular}[c]{@{}c@{}} \textbf{Human}\\  \textbf{Verified} \end{tabular}} } & \multicolumn{2}{c}{{\begin{tabular}[c]{@{}c@{}} \textbf{Overall}\\  \textbf{Test} \end{tabular}} } \\ \cline{2-7} 
 & \textbf{EM} & \textbf{F1} & \textbf{EM} & \textbf{F1} & \textbf{EM} & \textbf{F1}  \\ \hline
 {\begin{tabular}[c]{@{}c@{}} NQG\\  \end{tabular}}  & 66.99 & 79.67 & 64.71 & 79.36 & 65.26 & 79.43  \\\hline
 + QPP & \textbf{74.36} & \textbf{85.18} & \textbf{68.82} & \textbf{82.89} & \textbf{70.09} & \textbf{83.44} \\ \hline
+ Top-k & 71.58 & 83.48 & 66.77 & 80.45 & 67.94 & 81.19  \\
+ Tok-k + QPP & 72.52 & 84.98 & 67.67 & 81.79 & 68.84 & 82.56 \\ \hline
+ Nucleus & 70.62 & 83.68 & 67.16 & 80.37 & 68.00 & 81.17 \\ 
+ Nucleus + QPP & 74.12 & 85.08 & 68.10 & 81.36 & 69.56 & 82.26 \\

\hline\hline
 \end{tabular}%
}
\end{table}

\noindent\textbf{Performance of QPP with Sampling Strategies.}
Since our QPP is compatible with sampling strategies, we further study the performance after combining these two techniques. Table \ref{tbl:qa_results_side} shows the results, which indicate that combining two techniques can improve the sampling strategies' performance but do not lead to further improvement compared with using QPP only. This demonstrate that our QPP module is good enough to generate diverse useful questions for improving QA.



\begin{table}[t]
\centering
\caption{Choosing seq2seq-based QPP over alternative multi-label classification methods. HL: Hamming Loss.}
\begin{tabular}{lcccc}
\hline
Models & \textbf{HL} & \textbf{Precision} & \textbf{Recall} & \textbf{F1} \\ \hline
Binary Relevance & 0.0524 & \textbf{99.22} & 90.89 & 94.87 \\
Classifier Chain & 0.0524 & \textbf{99.22} & 90.89 & 94.87 \\ \hline
\textbf{QPP} & \multicolumn{1}{l}{\textbf{0.0346}} & \multicolumn{1}{c}{97.28} & \multicolumn{1}{l}{\textbf{96.20}} & \multicolumn{1}{l}{\textbf{96.74}} \\ 
\hline
\end{tabular}
\label{tbl:multi_label}
\end{table}


\noindent\textbf{Alternative Approaches for QPP.} There are many model options for the QPP task, e.g., those for multi-label classification. To justify our choice of a seq2seq model, we compare it with two commonly-adopted multi-label classification methods: binary relevance (BR) and classifier chain (CC) \cite{boutell_luo_shen_brown_2004,read2011classifier}. BR develops multiple binary classifiers independently while CC builds a chain of classifiers and predicts labels sequentially. We use multi-layer perceptron as the base model for both BR and CC. For each answer evidence, the input is the representation from the same LSTM encoder as our QPP module.


From Table \ref{tbl:multi_label}, we can see: (1) The seq2seq design in our QPP module performs better overall and especially in terms of Recall, which is particularly important since we aim for generating diverse question types; (2) A simple seq2seq model achieves great performance across all metrics, which renders developing more complex models for this task less necessary.



\section{Conclusion}
This paper proposes a simple yet effective framework for improving clinical QA on new contexts. It leverages a seq2seq-based question phrase prediction module to enable QG models to generate diverse questions. Our comprehensive experiments and analyses allow for a better understanding of why diverse question generation can help QA on new clinical documents.

\section*{Acknowledgment}
The authors would like to thank all the constructive reviews. The research is sponsored in part by the PCORI Funding ME-2017C1-6413, the Army Research Office under cooperative agreements W911NF-17-1-0412, NSF Grant IIS1815674, NSF CAREER \#1942980, and Ohio Supercomputer Center \cite{OhioSupercomputerCenter1987}.
The views and conclusions contained herein are those of the authors and should not be interpreted as representing the official policies,
either expressed or implied, of the Army Research Office or the U.S. Government. The U.S. Government is authorized to reproduce and distribute reprints for Government purposes notwithstanding any copyright notice herein.




\end{document}